\documentclass[letterpaper]{article} 
\usepackage{aaai25}  
\usepackage{times}  
\usepackage{helvet}  
\usepackage{courier}  
\usepackage[hyphens]{url}  
\usepackage{graphicx} 
\usepackage{natbib}  
\usepackage{caption} 
\usepackage{algorithm}
\usepackage{algorithmic}
\usepackage{amssymb}
\usepackage{amsmath}
\usepackage{cases}
\usepackage{multirow}

\usepackage{newfloat}
\usepackage{listings}
\DeclareCaptionStyle{ruled}{labelfont=normalfont,labelsep=colon,strut=off} 
\floatstyle{ruled}
\newfloat{listing}{tb}{lst}{}
\floatname{listing}{Listing}
\title{$\begin{array}{l}\includegraphics[height=2.2\fontcharht\font`\B]{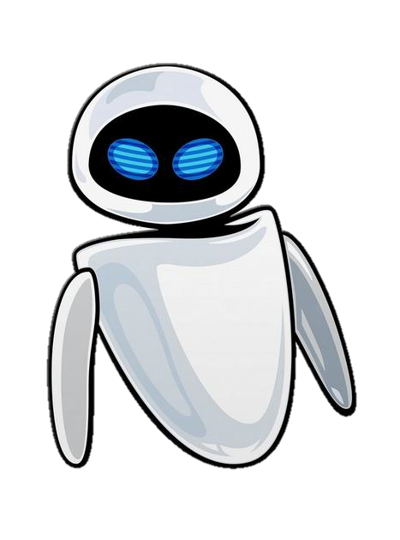}\end{array}$\textbf{Eve}: Efficient Multimodal Vision Language Models with Elastic Visual Experts}
\author{
    Miao Rang,
    Zhenni Bi,
    Chuanjian Liu,
    Yehui Tang,
    Kai Han\thanks{Corresponding Author.},
    Yunhe Wang$^{\ast}$
}
\affiliations{
    Huawei Noah’s Ark Lab\\

    \{rangmiao1,bizhenni, liuchuanjian, yehui.tang, kai.han,yunhe.wang\}@huawei.com}
\usepackage{bibentry}

\begin{document}

\maketitle

\begin{abstract}

Multimodal vision language models (VLMs) have made significant progress with the support of continuously increasing model sizes and data volumes. Running VLMs on edge devices has become a challenge for their widespread application. There are several efficient VLM efforts, but they often sacrifice linguistic capabilities to enhance multimodal abilities, or require extensive training. To address this quandary,  we introduce the innovative framework of Efficient Vision Language Models with Elastic Visual Experts (\textbf{Eve}). By strategically incorporating adaptable visual expertise at multiple stages of training, Eve strikes a balance between preserving linguistic abilities and augmenting multimodal capabilities. This balanced approach results in a versatile model with only \textbf{1.8B} parameters that delivers significant improvements in both multimodal and linguistic tasks. Notably, in configurations below 3B parameters, Eve distinctly outperforms in language benchmarks and achieves state-of-the-art results \textbf{68.87$\%$} in VLM Benchmarks. Additionally, its multimodal accuracy outstrips that of the larger 7B LLaVA-1.5 model. Our code is available at https://github.com/rangmiao/Eve.
\end{abstract}
\begin{figure*}
  \centering
 \includegraphics[width=0.9\linewidth]{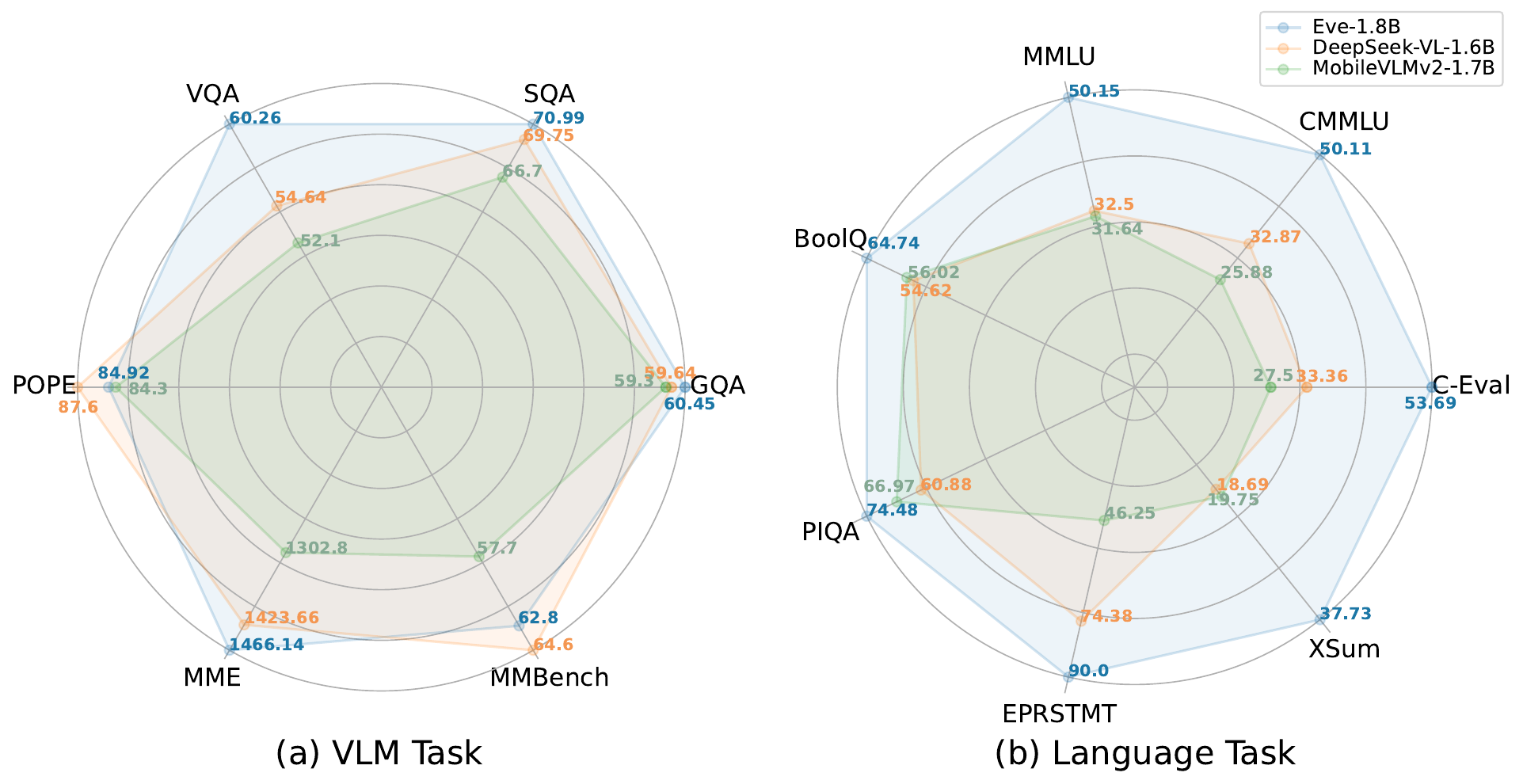} 
  \hfill
  \setlength{\abovecaptionskip}{-0cm}
  \setlength{\belowcaptionskip}{-0.5cm}
  \caption{Comparison with SOTA methods with 1B scale across VLM and language benchmarks.}
  \label{fig:wrapfig}
\end{figure*}

%

\section{Introduction}

As the swiftly evolving of artificial intelligence, the understanding of vision and language has gained significant attention, becoming a prominent research focus. Multi-modal models, such as Vision-Language Models (VLMs), are designed to combine visual information and textual descriptions, aiming to enhance semantic comprehension. These models, including GPT4V~\cite{gpt4v} and Gemini~\cite{team2023gemini}, have shown substantial potential in various applications, such as visual reasoning, visual question answering, and multi-modal retrieval. 

Most of the existing VLMs primarily enhance multimodal capabilities by expanding data volumes or enlarging the model sizes. Consequently, numerous high-quality visual-textual datasets~\cite{zhang2024llavar,chen2023sharegpt4v,zhao2023svit,chen2024allava} have been developed alongside a suite of large model tuning techniques~\cite{team2023gemini,bai2023qwenvl,liu2023improved,liu2024visual}. These approaches not only boost the model's generalization capabilities, enabling it to adeptly handle a diverse range of visual and textual inputs, but also enhance its ability to recognize and comprehend complex real-world scenarios and relationships. However, these models are usually large in size, making it difficult to deploy and perform efficient inference on the devices, hindering their practical applications.

To develop efficient VLMs, several methods are proposed to maintain multimodal capabilities while reducing the model size~\cite{chu2023mobilevlm,lin2024moellava,yuan2024tinygptv}. However, these methods often focus on augmenting multimodal capabilities at the expense of linguistic proficiency. MoE-LLAVA~\cite{lin2024moellava}, for instance, significantly enhances multimodal capacities by integrating multiple experts. However, a notable degradation in the precision of language tasks. Conversely, DeepSeek-VL~\cite{lu2024deepseek} maintains linguistic abilities during multimodal training by incorporating substantial amounts of language data. Although effective, this strategy heavily enlarges the training cost.

In this paper, we propose an efficient VLM framework to build multimodal and language capabilities under relatively small model size and low training cost. Based on the existing powerful LLMs, we introduce elastic vision experts to process visual inputs and enhance multimodal capabilities. The proposed Eve framework consists of three training stages and strategically embeds elastic vision experts at each stage. In the initial two stages, we can leverage the well pretrained vision encoders such as ResNet and Vision Transformer (ViT), in the public community which is elastic for building strong visual capability. In the third stage, we integrate an elastic vision feed forward network (FFN) in the LLM transformer while keeping the original LLM part frozen. This structured integration allows each expert to concentrate on distinct visual tasks, thereby enhancing multimodal capabilities without compromising the intrinsic linguistic abilities. Furthermore, this approach obviates the need for substantial textual data during training, thus significantly expediting the model training process. Within a model size of 3B parameters, Eve achieves state-of-the-art performance, and compared to other multimodal models, our language capabilities are better preserved.

In summary, the contributions of this paper are as follows:
\begin{itemize}
 \item 
 We present the Elastic Visual Expert (\textbf{Eve}) framework, meticulously structured across three training stages, and ingeniously incorporates dynamically adaptive visual experts in each phase, enabling each expert to concentrate on distinct domain-specific tasks. Throughout the training process, we strategically amalgamate the peak performance of these experts to bolster multimodal capabilities, all while maintaining the inherent linguistic proficiency;
\item 
The Elastic Visual Experts, featuring the Elastic Vision Encoder (EVE) and Elastic Vision Feed-Forward Network (EVF), are engineered with remarkable adaptability. During the first two stages of training, the visual encoder remains frozen, facilitating seamless integration with various visual encoders, while preserving the language model's performance. In the third stage, the EVF is introduced, unifying with the model's linguistic capabilities to create a powerful synergy. This fusion significantly elevates the model's ability to process and merge visual and textual data, thereby markedly enhancing its multi-modal performance;
\item 

Eve stands out in multimodal tasks with less than 3 billion parameters, achieving top performance in VLM and language benchmarks, and is on par with the larger 7B LLaVA-1.5 in terms of multimodal accuracy.
\end{itemize}
\section{Related Work}

\paragraph{Large Vision Language Models.}
As the capabilities of Large Language Models (LLMs) have significantly increased in tasks such as reasoning, comprehension, and question answering, Large Vision Language Models (LVLMs) are integrating powerful large language models with visual branches to expand the reasoning abilities of LLMs for processing multimodal data, thereby achieving more comprehensive and in-depth understanding and generation capabilities. In the field of visual-language learning, a notable example is CLIP~\cite{radford2021learning}, which employs a large number of image-text pairs for contrastive learning to align images and language in a semantic space. Building upon CLIP, BLIP~\cite{li2022blip} utilizes single-modal encoders for image and text encoding, with the text encoder, similar to BERT~\cite{devlin2019bert}, adding a new token [CLS] to represent the entire sentence in the input. BLIP-2~\cite{li2023blip2} introduces Q-Former to align the frozen visual base model and LLM. Additionally, MiniGPT-4~\cite{zhu2023minigpt4} introduce visual instruction fine-tuning through a projection layer, aligning a frozen visual encoder with an advanced frozen LLM Vicuna to enhance instruction following capabilities. ShareGPT4V~\cite{chen2023sharegpt4v} has generated a high-quality image-text description dataset covering a wide range of domains, including world knowledge, object properties, spatial relationships, and aesthetic evaluation, significantly improving the model's accuracy in multimodal benchmark testing. Qwen-VL~\cite{bai2023qwenvl} integrates training data from tasks such as image captioning, visual question answering, OCR, and document understanding, incorporating visual foundational capabilities into Qwen-VL. The generated model demonstrates outstanding performance across these diverse tasks.

\paragraph{Efficient Vision Language Models.}
The practical application of multimodal large language models has been limited by the computational costs and memory requirements during both training and inference stages. Recently, several studies have delved into the exploration of Small Vision Language Models (SVLMs) are characterized by a parameter spectrum that encompasses a range from 1 billion to 3 billion from different perspectives. For instance, LLaVA-Phi~\cite{zhu2024llavaphi} leverages the pre-trained small language model Phi-2 (2.7B) as the core of the multimodal model, incorporates CLIP ViT-L/14 as the visual encoder, and employs two layers of MLP to connect the visual encoder, demonstrating outstanding performance in visual understanding, reasoning, and multimodal perception. MobileVLM~\cite{chu2023mobilevlm, chu2024mobilevlm} provides an open-source approach for 1B/3B visual language models, enhancing the performance of SVLM through innovative adapter designs and high-quality data. TinyGPT-V~\cite{yuan2024tinygptv} utilizes economically efficient and powerful small language models to achieve robust and easily deployable models, suitable for various real-world visual language applications.

\paragraph{Mixture of Experts in Multi-modal Learning.}
The concept of Mixture of Experts (MoE) was first introduced in ~\cite{6797059} as a novel supervised learning process, which employs multiple models (or "experts") to learn and utilizes a gating network to determine which model is best suited to train on each data point. This approach reduces interference between different types of samples, enabling each expert to focus on processing a single task more effectively.
In V-MoE~\cite{riquelme2021scaling}, the authors introduced the first large-scale application of MoE to the Vision Transformer (ViT), significantly increasing accuracy while reducing computational costs. The VL-MoE~\cite{shen2023scaling} is the first work to apply MoE in the fusion of image and text modalities, demonstrating outstanding performance across multiple tasks. The VLMo~\cite{bao2022vlmo} model employs three experts, specializing in visual, linguistic, and visual-linguistic tasks, using the MoE framework to balance the depth encoding of each modality and the fusion of multimodal information. This design enables flexible handling of both single-modal and multimodal data pairs.
Building upon the VLMo model structure, BEiT-3~\cite{wang2022image} further expands the model's scale and simplifies the pre-training loss function. MoE-LLaVA~\cite{lin2024moellava} proposes a sparse model architecture based on MoE, featuring a soft router, which requires fewer activation parameters to achieve or even surpass the performance of dense models.

\section{Methods}

\subsection{Overview}

\begin{figure*}
  \centering
 \includegraphics[width=0.98\linewidth] 
            {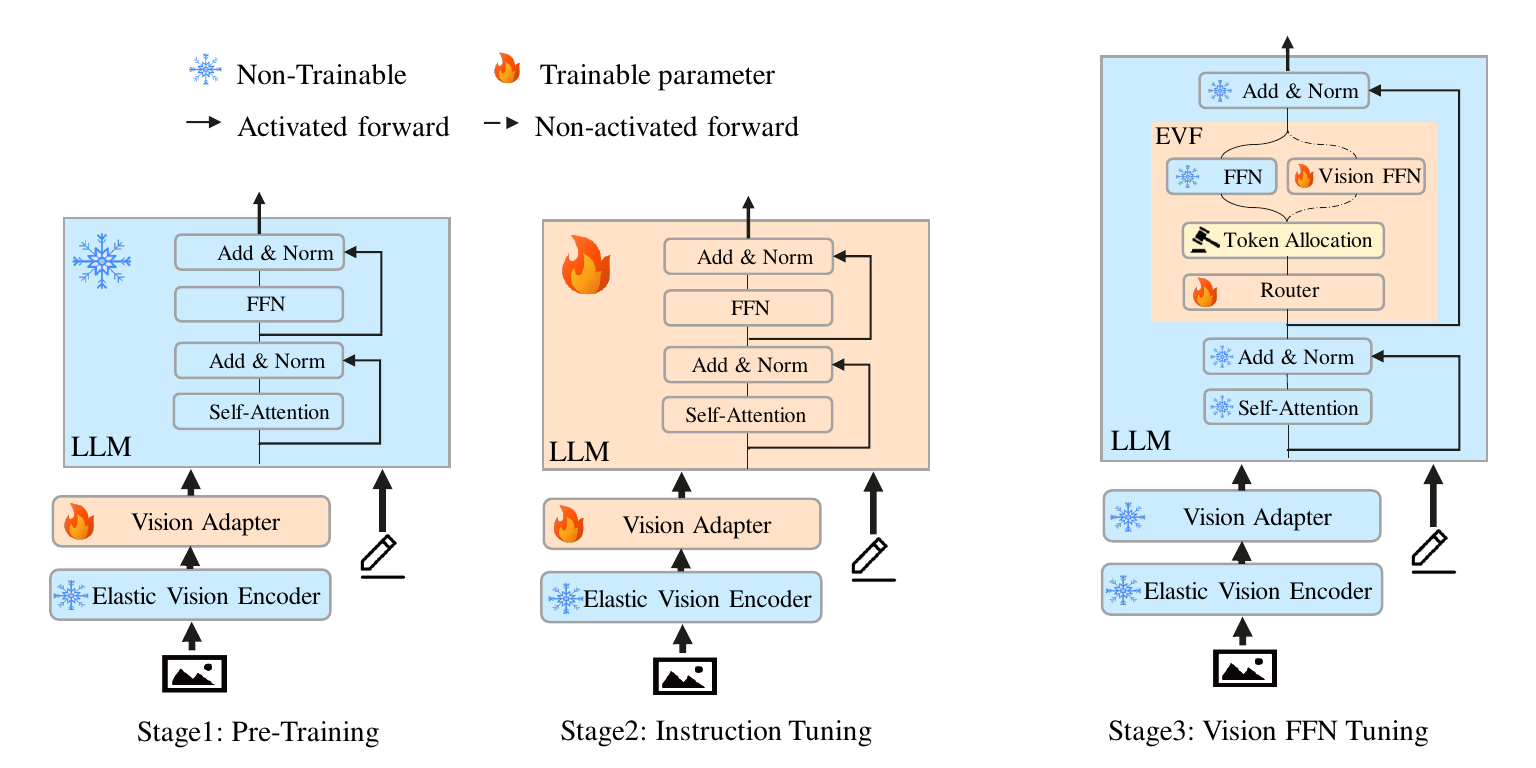} 

  \hfill
  \setlength{\abovecaptionskip}{0cm}
  \setlength{\belowcaptionskip}{-0.4cm}
  \caption{The Eve training framework and strategy. The Eve employs a meticulously structured three-stage training approach. \textbf{Stage 1}: Training is dedicated to the vision adapter to adapt the LLM specifically for processing visual inputs. \textbf{Stage 2}: To enhance multimodal capabilities, training vision adapter and LMM with LoRA. \textbf{Stage 3}: We introduce a new EVF layer, consisting of an elastic vision FFN and a fixed language FFN. The weights from the original FFN are duplicated to initialize the two FFNs in alternating half-layers of the LLM. This stage involves isolated training of the vision FFN, aimed at significantly enhancing the model's proficiency in visual information comprehension.}
  \label{fig:structure}
\end{figure*}

Our proposed model, \textbf{Eve}, incorporates a sophisticated three-stage framework, strategically integrating elastic vision experts at each stage, as depicted in Fig.~\ref{fig:structure}. A key focus of our approach is the preservation of linguistic capabilities throughout the training process. Notably, the linguistic proficiency of the model remains unaffected by the variations in pre-training data used for the visual encoder during the first two stages of training. This stability in linguistic performance is a significant accomplishment, as it ensures that the model's ability to process and comprehend language is not compromised by changes in the visual encoder's pre-training. Further details on this aspect are provided in Elastic Vision Encoder.


In the later stages of the training strategy, we introduce a novel elastic visual FFN in the third phase to enhance the model’s capacity for multimodal data processing. This component is specifically designed to complement the model's linguistic capabilities, allowing it to effectively analyze visual data while preserving its proficiency in language tasks. The integration of the elastic visual FFN not only strengthens the model's multimodal abilities but also ensures the retention of its inherent linguistic competencies. As a result, the model is capable of handling complex multimodal inputs while preserving its linguistic processing and comprehension capabilities. A detailed discussion of the design and implementation of the elastic visual expert is provided in Elastic Vision FFN.
\subsection{Elastic Vision Encoder}

Over the past decade, the research community has developed numerous foundational visual models, such as ResNet~\cite{he2016deep} and ViT~\cite{dosovitskiy2020image}, which exhibit exceptional capabilities in visual signal processing and are capable of extracting rich visual representations. Building upon these advancements, we propose an elastic vision encoder that harnesses the strengths of existing visual models, effectively standing on the shoulders of giants to construct a more robust vision-language model (VLM).

\paragraph{Leveraging Vision Encoders in the Wild.}
The vision encoder extracts features from an RGB image, fundamentally transforming this input into a sequence of visual embeddings that capture critical visual feature details. To enhance multimodal capabilities while preserving the inherent abilities of the language model, the vision adapter is continuously trained to align vision features with the language model's feature space in the first two stages, while the language model undergoes only light LoRA-based~\cite{hu2021lora} fine-tuning in the second stage, as shown in Fig \ref{fig:structure}. The vision encoder remains frozen to elastically support any vision backbone models. This design ensures that the language model's performance remains largely unaffected even when the vision encoder is flexibly replaced. Consequently, our architecture can incorporate elastic vision encoders, including the use of various pre-training data. There are a large number of pretrained vision backbones in the opensource community, which elastically provide a rich source of vision encoders. Therefore our approach can maximize the use of existing industry capabilities to enhance multimodal capabilities while preserving the inherent linguistic abilities of the language model. 

\paragraph{Pre-training Data in Vision Encoder.}

To minimize the temporal costs associated with individual trials, our study employs smaller-scale visual models in conjunction with the PanGu-$\pi$-1.5B~\cite{wang2023pangu} language model. Specifically, we utilize the ResNet-50 architecture~\cite{he2016deep} as the vision encoder, evaluating its performance across different datasets: ImageNet-1K~\cite{russakovsky2015imagenet}, ImageNet-22K~\cite{ridnik2021imagenet}, and LAION-400M~\cite{radford2021learning}. The experimental results are summarized in Table~\ref{data-vision}. We observe that the model trained on ImageNet-22K achieves the highest precision, with a value of 53.36$\%$ on the VLM benchmark. Although LAION-400M, being a larger dataset, contains more diverse internet-sourced data, its quality is compromised by the presence of non-ideal samples. In contrast, ImageNet-22K offers higher-quality data, demonstrating that superior data quality can outweigh the benefits of larger, lower-quality datasets in training visual models. Notably, the accuracy difference between the best-performing model (trained on ImageNet-22K) and the lowest-performing one (trained on LAION-400M) is within a narrow 0.3$\%$ range, as shown in the final column of Table~\ref{data-vision}. Furthermore, pre-training the vision encoder with elastic pre-trained datasets effectively preserves the linguistic capabilities of the model.

\begin{table}[t]

  \setlength{\abovecaptionskip}{0cm}
  \setlength{\belowcaptionskip}{-0.5cm}
	\begin{center}
        \fontsize{9}{9}\selectfont
        \setlength{\tabcolsep}{0.5mm}
        \renewcommand{\arraystretch}{1.2}
            {
    		\begin{tabular}{c|c|c|c|c}
         
    			\hline
    			\bf{Vision Encoder}&\bf{P-Dataset} & \bf{Data Size} &\bf{VLM AVG}& \bf{L-AVG} \\
    			\hline
                ResNet50 & ImageNet-1K & 1.2M  & 49.06 & 51.73\\
    			ResNet50  & ImageNet-22K & 14M & \textbf{53.36} & \textbf{51.98}  \\
    			ResNet50  & LAION-400M & 413M & 52.68 & 51.68 \\
    			\hline
    		\end{tabular}}
	\end{center}
    \caption{Vision encoder pre-training dataset impact on VLM and language tasks. P-Dataset  denotes the pre-training dataset utilized for the vision encoder, and L-AVG is short for Language Task Average Accuracy.}
    \label{data-vision}
\end{table}

\subsection{Elastic Vision FFN}

In the third phase, inspired by MoE-LLaVA~\cite{jiang2024mixtral}, we introduce an Elastic Vision Feed-Forward Network (EVF) to enhance the multimodal capabilities of the model. This addition is designed to improve the processing of visual information within the large language model (LLM). To preserve the model’s existing linguistic proficiency, we freeze the majority of the parameters in the language model, allowing only the parameters of the EVF layer to be updated during training.

\paragraph{Elastic Vision FFN Layer (EVF).} 

As depicted on the right side of Fig.~\ref{fig:structure}, each EVF layer incorporates a sophisticated routing mechanism and a dedicated token allocation strategy, alongside two distinct Feedforward Networks (FFNs): one specialized for linguistic processing and the other for visual information. This dual-FFN design significantly enhances the model's multimodal capabilities by efficiently processing both language and visual data.



During forward propagation in the large language model (LLM), image tokens processed through the vision adapter and text tokens are concatenated and jointly fed into the LLM. Initially, the routing layer assigns each token to a recommended Feedforward Network (FFN). The token allocation mechanism then considers both the routing layer’s recommendation and the current capacity of the FFN to determine whether the token should be assigned to that specific FFN. The routing mechanism employs a linear layer to compute the probabilities of assigning each token to its respective FFN, selecting the one with the highest probability as the preferred FFN. The routing mechanism can be formalized as follows:

\vspace{-0.5cm}
\begin{center}
\begin{equation}
\label{eq4}
    	\mathrm{P}(\mathbf{x})_{i}=\frac{e^{f(\mathbf{x})_{i}}}{e^{f(\mathbf{x})_{l}} + e^{f(\mathbf{x})_{v}}},
\end{equation}
\end{center}


where the router generates weight logits $f(x) = W \cdot x$, which are then normalized using the softmax function. Here, $W$ represents the lightweight trainable parameters. The logit $f(x)_{l}$ corresponds to the language FFN, while $f(x)_{v}$ corresponds to the vision FFN.


In the initialization phase of Stage 3, we duplicate the FFN parameters from Stage 2 into both the language and vision FFNs. During the training phase of Stage 3, we freeze all parameters of the vision encoder and vision adapter, as well as the majority of the parameters in the language model, restricting training to the vision FFN and the routing layer within the language model. In the inference phase for multimodal tasks, both FFNs within the EVF layer are activated. The routing and token allocation mechanisms collaborate to assign tokens to the appropriate FFN. For language-only tasks, however, the EVF layer excludes both the routing layer and the vision FFN, retaining only the untrained language FFN. As a result, the language model operates in its standard configuration, with the language FFN remaining untrained and its linguistic capabilities fully preserved.

\paragraph{Token Allocation.} 


The token allocation mechanism plays a pivotal role in determining which Feedforward Network (FFN) each token is assigned to. In the EVF layer, each FFN $e_{i}$ has a predefined capacity ${C}$, which limits the number of tokens it can process. In conventional token allocation mechanisms, if the number of tokens ${M}$ recommended by the routing mechanism for an FFN $e_{i}$ exceeds its capacity ${C}$, only a random subset of ${C}$ tokens is selected from ${M}$ for allocation to FFN $e_{i}$, and the excess ${M}$-${C}$ tokens are discarded. This indiscriminate discarding of tokens can significantly degrade the model's accuracy.


To overcome the limitations of conventional token allocation, we introduce GBPR, a novel strategy that prioritizes token distribution within a complete batch based on token importance, as determined by the routing score ${P}(x)$. Specifically, when the number of tokens ${M}$ exceeds the capacity ${C}$ of an FFN $e_{i}$, GBPR prioritizes the allocation of the ${C}$ most important tokens to $e_{i}$, while discarding the remaining less important ${M}-{C}$ tokens.


Further improving this approach, we propose Img-GBPR, a mechanism for distinctly managing vision and text tokens. This mechanism assigns a default recommended FFN for each token type to ensure that visual and linguistic FFNs can focus on their respective tasks. Image tokens are initially assigned a score $S_i \in \mathbb{R}^{P \times 2}$, with values in the visual FFN column approaching 1, thus directing them primarily to the visual FFN. In contrast, values in the linguistic FFN column are near 0. Text tokens are scored differently, with $S_t \in \mathbb{R}^{N \times 2}$, where the score for the visual FFN approaches 0 and the score for the linguistic FFN approaches 1, facilitating their allocation to the linguistic FFN. The final priority score for each token is calculated by summing the routing score ${P}(x)$ and the initial score $S_i$ or $S_t$, optimizing token allocation based on their modality. The score can be represented as follows:

\begin{center}
\begin{equation}
\label{eq1}
    S(x)=\begin{cases}
  {P}(x) + S_{i}, & \text{  if } x \in \text{image token} \\
  {P}(x) + S_{t}, & \text{ if } x \in \text{text token} ,
\end{cases}
\end{equation}
\end{center}
Furthermore, when the number of tokens ${M}$ assigned to a FFN $e_{i}$ exceeds its capacity ${C}$, we prioritize the selection of the most important ${C}$ tokens based on ${S}(x)$ for allocation to FNN $e_{i}$. The remaining ${M}-{C}$ unallocated tokens are reintroduced to the candidate pool for redistribution. A certain proportion ${W}$ is randomly selected and allocated to another FFN to minimize the loss of tokens. This scoring strategy ensures that image tokens are preferentially assigned to the visual FFN, optimizing their processing for visual tasks, while text tokens are directed to the linguistic FFN, thereby enhancing the model’s capabilities for textual interpretation. This dual-token allocation approach maximizes the efficiency of each FFN, enabling them to operate optimally within their respective modalities.

\paragraph{Balanced Loss.} 

We draw upon MoE-LLaVA, where the overall loss function consists of both the regression loss 
$\mathcal{L}_{\text {regressive }}$ and the auxiliary loss $\mathcal{L}_{\text {aux}} $. The regression loss optimizes the model's performance, while the auxiliary loss encourages balanced load distribution across the FFNs. Given our unique token allocation mechanism, we have adjusted the coefficient $\alpha$  of the auxiliary loss to 0.001.

\vspace{-0.6cm}
\begin{center}
\begin{equation}
\label{eq4}
\mathcal{L}_{\text {total }}=\mathcal{L}_{\text {regressive }}+\alpha \cdot \mathcal{L}_{\text {aux }},
\end{equation}
\end{center}
We integrate a differentiable load balancing loss into each EVF layer to promote equitable distribution of token processing among FFN, as follows:
\vspace{-0.4cm}
\begin{center}
\begin{equation}
\label{eq4}
\mathcal{L}_{\text {aux }}={F}_{i} \cdot {G}_{i} + {F}_{t} \cdot {G}_{t}, 
\end{equation}
\end{center}

where ${F}_{i}$ and ${F}_{t}$ denote the fraction of tokens processed by vision and language FFN, respectively, while ${G}_{i}$ and ${G}_{t}$ represent the average routing probabilities for vision and language FFN. 

\section{Experiments}
\subsection{Settings}
\paragraph{Model Details.} Eve is meticulously designed around three core components: a vision adapter, a visual encoder, and a language model. The vision adapter, empowered by the Lightweight Downsample Projector (LDP)\cite{chu2023mobilevlm}, serves as an innovative bridge between the visual encoder and the language model, enabling seamless integration and alignment of multimodal features. The visual encoder leverages SigLip-L\cite{zhai2023sigmoid}, built upon the robust VIT-L backbone and utilizing a patch size of 576, which is renowned for its exceptional ability to capture intricate details from visual data. Complementarily, the linguistic backbone is formed by is PanGu-$\pi$-1.5B-Pro~\cite{tang2024rethinking}, a powerful architecture featuring 22 layers, a width of 2048 dimensions, and a vocabulary size of 48,000 entries. This high-capacity design significantly enhances Eve’s capacity to comprehend nuanced language structures and generate sophisticated text, thereby strengthening its overall competence in cross-modal understanding and expression.

\paragraph{Implementation Details.} In Stage 1, we freeze both the vision encoder and the LLM, focusing exclusively on training the efficient vision adapter. In Stage 2, we fine-tune the vision adapter in conjunction with the LLM, utilizing the LoRA technique. Finally, in Stage 3, we train only the vision FFN, assigning each FFN a capacity of ${C}$=1.5. Detailed training settings are provided in Table~\ref{train_detail}.

\begin{table}[t]
	\begin{center}
  \setlength{\belowcaptionskip}{-0.5cm}

        \fontsize{8}{8}\selectfont
        \setlength{\tabcolsep}{0.5mm}
        \renewcommand{\arraystretch}{1.2}
            {
    		\begin{tabular}{c|ccc}
    			\hline
    			{\bf{Configuration}} & \bf{Stage 1} & {\bf{Stage 2}}& {\bf{Stage 3}} \\
                \hline
                Vision encoder init & SigLip-L & SigLip-L  & SigLip-L \\
                LLM init & PanGu-$\pi$-1.5B Pro & PanGu-$\pi$-1.5B Pro  & Eve Stage2 \\
                Vision adapter init & Random & Eve Stage1  & Eve Stage2 \\
                Image resolution & 384x384 & 384x384 & 384x384 \\
                Learning rate & 1e-3 & 2e-5 & 2e-5 \\
                LR schedule & Cosine decay & Cosine decay & Cosine decay \\
                Weight decay & 0 & 0 & 0 \\
                Optimizer & \multicolumn{3}{c}{AdamW($b_{1}$=0.9, $b_{2}$=0.95)} \\
                Warmup ratio & 0.03 & 0.03 & 0.03 \\
                Global batch size & 256 & 128& 128 \\
                Training steps & 2181 & 5197 & 25510  \\
                Training hours & 0.8 & 7 & 34   \\
                Epoch & 1 & 1& 1 \\
                GPU &  8xV100-32G & 8xV100-32G & 8xV100-32G \\
                \hline
    		\end{tabular}}
	\end{center}
	%
	\caption{Training hyperparameters of Eve.}
	\label{train_detail}
\end{table}

\paragraph{Training Dataset.}Our dataset has been carefully refined and expanded to create high-quality datasets that enhance cross-modal understanding. In the first two phases, we utilize the CC-595K and LLaVA-mixed-665 datasets to develop foundational multimodal capabilities. In the third phase, we curate a diverse collection of datasets across several domains, including General Multi-modality, Visual Question Answering (VQA), Optical Character Recognition (OCR), Image Captioning, and Knowledge-intensive tasks. This comprehensive ensemble consists of over 3.2 million samples, all meticulously designed to significantly enhance the model's versatility and performance across a wide range of modal scenarios. Detailed descriptions of the various datasets are provided in Appendix B. 


 %
\paragraph{Evaluation.}  Our primary objective centers on rigorously evaluating the model's proficiency in both multimodal and linguistic tasks. Following the rigorous evaluation protocols established in prior works such as \cite{chu2023mobilevlm, chu2024mobilevlm}, we employ a comprehensive suite of VLM benchmarks for multimodal assessment, comprising GQA~\cite{hudson2019gqa}, SQA~\cite{lu2022learn}, TextVQA~\cite{singh2019towards}, MME~\cite{fu2023mme}, MMBench~\cite{liu2023mmbench} and POPE~\cite{li2023evaluating}. Consistent with the approach outlined in~\cite{tang2024rethinking}, we employ a diverse array of benchmarks to evaluate linguistic competencies. These include  C-Eval~\cite{huang2024c}, CMMLU~\cite{li2023cmmlu}, MMLU~\cite{hendrycks2020measuring}, BoolQ~\cite{clark2019boolq}, PIQA~\cite{bisk2020piqa}, EPRSTM~\cite{xu2021fewclue} and XSum~\cite{narayan2018don}.

\subsection{Ablation Study}
\subsubsection{Effect of Elastic Vision FFN Layers.}


In the third stage, we compare the performance differences between the EVF and MoE layers on multimodal and language tasks, using a ResNet50 vision encoder and PanGu-$\pi$-1.5B as the language model. The MoE layers, introduced by MoE-LLAva, are adjusted to match the parameter count of the EVF layers by employing a "x2top1" strategy, which involves two FFNs without differentiation; both FFNs are activated and trained simultaneously. Detailed comparisons are presented in Table~\ref{EVE-layer}. Compared to Stage 2, the MoE layers improve multimodal task accuracy by 0.55$\%$, but significantly reduce language task accuracy by 3$\%$. In contrast, the EVF layer architecture not only enhances multimodal task accuracy by 0.47$\%$, but also fully preserves language task accuracy.

\begin{table}[ht!]
	\begin{center}
            {
        \renewcommand{\arraystretch}{1.2}
    		\begin{tabular}{l|cccc}
    			\hline
    			 {\bf{Stage}} & {\bf{EVF}} & {\bf{MoE}} & {\bf{VLM AVG}} & {\bf{Language AVG}}\\

    			\hline
              Stage2 & &  & 61.23 &  58.65 \\
    	   Stage3 &  & \checkmark &  \textbf{62.23} &  55.03 \\ 
    		 Stage3 & \checkmark  &  & 61.93 & \textbf{58.65} \\
                \hline
    		\end{tabular}}
	\end{center}
	\caption{Impact of EVF vs. MoE Layers on VLM and Language benchmark}
	\label{EVE-layer}
\end{table}

\begin{table}[ht!]
  \setlength{\belowcaptionskip}{-0.7cm}
	\begin{center}
    \renewcommand{\arraystretch}{1.2}
    \begin{tabular}{c|c}
            \hline
          \bf{Token Allocation}  & {\bf{VLM AVG}} \ \\
                \hline

         Random & 53.83 \\
          GBPR & 54.37 \\
        Img-GBPR  & \bf{54.92} \\ 
          
            \hline
        \end{tabular}
     \end{center}
    \captionof{table} {Imapct of different token allocation.}
    \label{token_sellection_effect}
\end{table}

\begin{figure*}
  \centering
 \includegraphics[width=0.98\linewidth] 
            {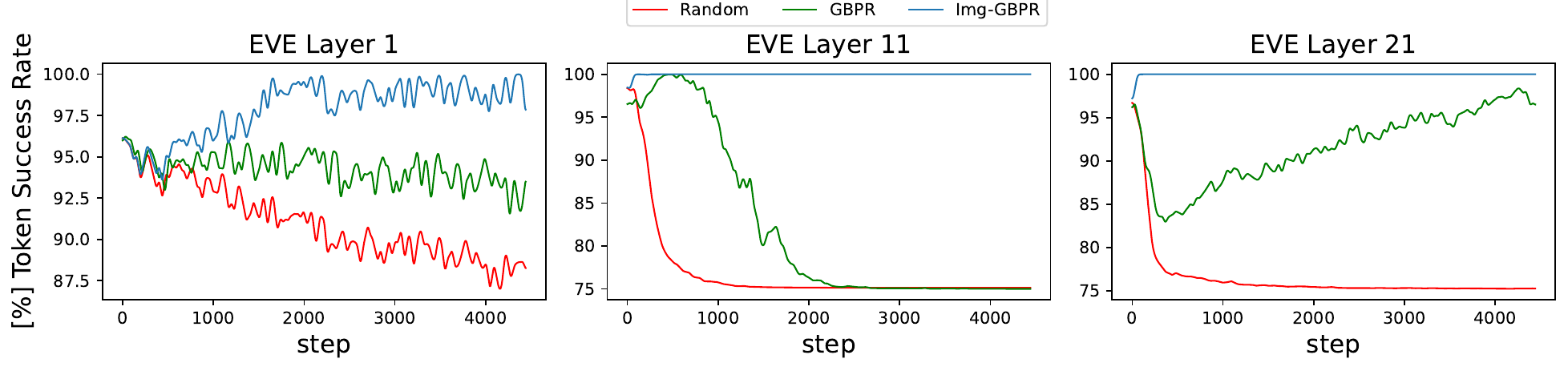} 
  \hfill
	\caption{The impact of token allocation mechanisms on successful routing in Layer 1, 11 and 21.}
	\label{fig.bpr_2}
\end{figure*}

\begin{table*}[ht]
	\begin{center}
         \fontsize{8}{8}\selectfont
        \setlength{\tabcolsep}{0.8mm}
        \renewcommand{\arraystretch}{1.2}
            {
    		\begin{tabular}{l|cccc|ccc|ccc|c|c}
                    \hline
    			   & & & & &\multicolumn{3}{c|}{\bf{Image Question Answering}}     &\multicolumn{3}{c|}{\bf{Benchmark Toolkit}} & \ \\
    			\cline{6-11}
    			{\bf{Methold}} & {\bf{VM}} & {\bf{LLM}} & {\bf{Act.}}  & {\bf{Res.}} & {\bf{GQA}}& {\bf{SQA$^\mathrm{I}$}} & {\bf{VQA$^\mathrm{T}$}}& {\bf{POPE}} & {\bf{MME$^\mathrm{P}$}}& {\bf{MMB$^\mathrm{dev}$}}& {\bf{AVG}}& {\bf{GPU-days}} \\
                & & & & &12578  & 2017 & 5000  & 8910  & 2374 & 4329 &  \\
       
    			\hline
                LLaVA-1.5~\cite{liu2023improved} & ViT-L & V-7B & 6.7B & 336 & 62.00 & 66.80 & 58.20 & 85.90 & 1510.70 & 64.30 & 68.79 & - \\
                LLaVA-1.6~\cite{liu2024llava} & ViT-L & V-7B & 6.7B & 336 & 64.80 & 72.80 & 65.70 & 86.70 & 1498.00 & 68.70 & 72.27 & - \\ 
                \hline
                TinyGPT-V~\cite{yuan2023tinygpt} & ViT-L & P-2.7B & 2.7B & 448 & 33.60 & 41.22 & 11.40 & 50.56 & 507.80 & 35.55 & 33.85 & - \\
                Mini-Gemini~\cite{li2024mini} & ConX-L & G-2B & 2B & 336 & - & - & 56.20 & - & 1341.00 & 59.80 & - & - \\
                
    		 MobileVLM~\cite{chu2023mobilevlm} & ViT-L & M-1.4B & 1.4B & 336 & 56.10  & 57.30 & 41.50 & 84.50 & 1196.20  & 53.20 & 58.70 & 1.6 \\
    		 MobileVLM~\cite{chu2023mobilevlm} & ViT-L & M-2.7B  & 2.7B & 336 & 58.40 & 59.00 & 46.70 & 84.60 & 1296.40  & 57.00 & 61.75 & 2.6\\
    	   MobileVLM v2~\cite{chu2024mobilevlm} &  ViT-L & M-1.4B & 1.4B & 336 &  59.30  & 66.70 & 52.10 & 84.30 & 1302.80  & 57.70 & 64.20 & 9 \\ 
    	   MobileVLM v2~\cite{chu2024mobilevlm} &  ViT-L & M-2.7B & 2.7B & 336 &  61.10  & 70.00 & 57.50 & 84.70 & 1440.50  & 63.20 & 68.10 & - \\ 
    	   LLaVA-Phi~\cite{zhu2024llavaphi} &  ViT-L & P-2.7B & 2.7B & 336 & \textbf{68.40} & 66.39 & 48.60 & 85.00 & 1335.10  & \textbf{66.70} & 65.82 & 3.2  \\        
    	   MoE-LLaVA-1.6Bx4-Top2~\cite{lin2024moellava} &  ViT-L & S-1.6B &  2.0B & 384 &  61.5  & 63.9 & 54.3 & 85.9 & 1335.7  & 63.3 & 65.95 & 8.6 \\ 
        
             DeepSeek-VL~\cite{lu2024deepseek} & SigLIP & D-1.3B &  1.3B &384 & 59.64 & 69.75 & 54.64 & \textbf{87.60} & 1423.66 & 64.60 & 67.90& 896 \\
    
                \hline 
             \textbf{Eve-VLM} & ViT-L & PG-1.5B & 1.5B &336 & 59.99 & 68.12 & 57.78 & 84.95 & 1292.35 & 60.82 & 66.05 & 15\\ 

             \textbf{Eve-VLM} & SigLIP & PG-1.5B &  1.5B &384 & 60.45 & \textbf{71.49} & \textbf{60.26} & 84.92 & \textbf{1466.14} & 62.80 & \textbf{68.87} & 15\\ 
             \hline
    		\end{tabular}}
	\end{center}
	\caption{Comparison with SOTA methods across 6 VLM benchmarks. 'VM' signifies the vision model component utilized in the VLM, whereas 'LLM' indicates the language model component.'Act.' refers to the number of activated parameters within the models,  'Res.' denotes input image resolution. The models 'V', 'M', 'P', 'S', 'D', 'G', and 'PG' correspond to Vicuna~\cite{chiang2023vicuna}, Mobile LLaMA~\cite{chu2023mobilevlm},  Phi-2~\cite{li2023pace}, StableLM~\cite{stabellm}, DeepSeek-LM~\cite{guo2024deepseek}, Gemini~\cite{li2024mini}, and PanGu-$\pi$-1.5B-Pro~\cite{tang2024rethinking}, respectively. "AVG" stands for the weighted mean of 6 VLM benchmarks. 'GPU-days' quantifies the computational time required for model training. }
	\label{sota_result_inVQA}
\end{table*}

\subsubsection{Effect of Token Allocation.}

We visualized the token success rate across different EVF layers during training using three distinct token allocation methods: Random, GBPR, and Img-GBPR. The analysis focused on layers 1, 11, and 21, with the results shown in Fig.~\ref{fig.bpr_2}. When employing the random token distribution mechanism, approximately 25$\%$ of tokens were discarded at layers 11 and 21. In contrast, GBPR improved the acceptance rates of tokens in the initial and final layers (Layer 1 and Layer 21) as training progressed, although a drop-off rate of approximately 25$\%$ persisted at layer 11 in the later stages. The introduction of Img-GBPR, with its redistribution strategy, resulted in a more substantial improvement in token success rates across all layers (initial, middle, and final), highlighting the effectiveness of token distribution strategies in optimizing model training.

Furthermore, we compared the impact of different token allocation methods on the accuracy of multimodal tasks, using a ResNet50 vision encoder and the PanGu-$\pi$-1.5B large language model. The experimental results are detailed in Table~\ref{token_sellection_effect}. Employing the GBPR method led to an improvement over the random allocation approach, with an average accuracy increase of 0.4 percentage points. When the Img-GBPR method was applied, model accuracy increased further by 0.5$\%$.

\subsubsection{Ablation Study of Best Results.}



To achieve optimal performance, we conduct a series of detailed ablation experiments across three dimensions: method, model, and training dataset. The specifics of these experiments are provided in Table~\ref{ablation_study_for_all}. The baseline model is based on MobileVLM, utilizing a ResNet50 visual encoder, MobileLLaMA as the language model, and LDP as the vision adapter, all trained on the Stage 2 dataset.

Initially, we replace the language model with PanGu-$\pi$-1.5B, resulting in a significant increase of 1.4$\%$ in average accuracy. We then incorporate two effective schemes that we propose—the EVF layer and Img-GBPR—which further improve accuracy by 1.6$\%$. To align with the current state-of-the-art model, DeepSeek, we replace both the visual and language models with stronger alternatives: the visual encoder is upgraded to SigLIP-L, which leads to a substantial 8$\%$ increase in multimodal accuracy. Additionally, replacing the language model with PanGu-$\pi$-1.5B-Pro further boosts accuracy by 1.5$\%$, reaching 64.52$\%$. Finally, substituting the training data with our meticulously curated Stage 3 dataset results in an additional 4.3$\%$ increase in accuracy, achieving a peak accuracy of 68.87$\%$.

\begin{table}[ht!]
	\begin{center}
       \setlength{\abovecaptionskip}{0.5cm}
      \setlength{\belowcaptionskip}{-1cm}
         \fontsize{9}{9}\selectfont
        \setlength{\tabcolsep}{0.8mm}
        \renewcommand{\arraystretch}{1.2}
            {
    		\begin{tabular}{lccccccc}
    			\hline
      		  \bf{ Vision}  & \multirow{2}{*}{\bf{LLM}} & \bf{EVF} & {\bf{Img}} & \bf{Stage3} & {\bf{VLM}} \ \\
       		  \bf{Encoder}  &  & \bf{Layers} &  \bf{GBPR}& \bf{Data} & {\bf{AVG}} \ \\
                    \hline
    
    			ResNet50 & MobileLLaMA & & & & 51.97 \\
                    ResNet50 & PanGu-$\pi$-1.5B  & & & & 53.36 \\
                    ResNet50 & PanGu-$\pi$-1.5B  & \checkmark& & &  53.83 \\
                    ResNet50 & PanGu-$\pi$-1.5B  & \checkmark & \checkmark &  & 54.92 \\
                    SigLip-L & PanGu-$\pi$-1.5B  & \checkmark &  \checkmark & & 63.03 \\
                    
                    SigLip-L & PanGu-$\pi$-1.5B-Pro  & \checkmark & \checkmark & & 64.52 \\
                    SigLip-L & PanGu-$\pi$-1.5B-Pro  & \checkmark & \checkmark & \checkmark & \textbf{68.87} \\
              
                \hline
    		\end{tabular}
      }
	\end{center}
	\caption{Ablation study for optimal results: effective methods, vision-language models, and training datasets."VLM AVG"  represents the average accuracy of the VLM benchmarks.}
	\label{ablation_study_for_all}
\end{table}

\subsection{Comparisons with State-of-the-art Methods}

We compare Eve with current state-of-the-art models in Table~\ref{sota_result_inVQA}. Among models with fewer than 3B activated parameters, Eve achieves the best accuracy 68.87$\%$. When compared to models with similar parameter sizes, Eve outperforms DeepSeek-VL by 1.9$\%$ and offers significant advantages in training efficiency, requiring only 15 GPU-days. Eve even surpasses that of some 7B models, such as LLaVA-1.5. Additionally, as shown in Fig.~\ref{fig:wrapfig}, Eve notably outperforms existing VLM with fewer than 3B parameters, especially in maintaining full language task capabilities. Detailed results are provided in Appendix A.3.

\section{Conclusion}


In this work, we introduce an efficient VLM framework, Eve, which embeds elastic visual experts at various stages. Additionally, the adaptive token allocation mechanism enhances the model's ability to process multimodal information effectively. As a result, the model not only retains its language capabilities but also significantly improves its multimodal performance.

\section{Acknowledgments}
We gratefully acknowledge the support of MindSpore, CANN (Compute Architecture for Neural Networks) and Ascend AI Processor used for this research.

\appendix
\section{Appendix}

\begin{table*}[ht!]

	\begin{center}

        \scalebox{0.98}
            {
            \renewcommand{\arraystretch}{1.2}
    		\begin{tabular}{l|c|c|c|c}
    			\hline
             \bf{Train Paradigm} & \bf{Vision Encoder} &  \bf{Pre-training Dataset} & \bf{VLM AVG} & \bf{Language AVG} \ \\
       
    			\hline
                    SL &  ResNet50  &  ImageNet-1K & 49.06 & 51.73 \\
                    SSL-MOCOv2 &  ResNet50  & ImageNet-1K & 50.34 & 51.92 \\
    			SSL-SWAV & ResNet50 & ImageNet-1K & 50.82 & 52.30 \\
    			SSL-DINO & ResNet50  & ImageNet-1K & \bf{50.95} & \bf{52.48} \\
    			\hline

    		\end{tabular}}
	\end{center}
 \caption{Performance of vision encoder pre-trained on diverse methodology in VLM and language benchmarks. "SL" denotes supervised learning, "SSL" denotes self-supervised Learning.}
 \label{Methodology-vision-language-task}
	
\end{table*}

\subsection{Pre-training Method in Vision Encoder}
To minimize the time cost associated with individual trials, our study experimented with the ResNet-50 architecture pre-trained on ImageNet-1K as the visual encoder and combined it with the language model PanGu-$\pi$-1.5B. We assessed the performance of this setup after training with methods including self-supervised, unsupervised, and supervised training. The self-supervised category included three variants: MoCo v2~\cite{chen2003improved}, DINO~\cite{caron2021emerging}, and SwAV~\cite{caron2020unsupervised}. Detailed results are presented in Table ~\ref{Methodology-vision-language-task}. All three self-supervised techniques exhibited higher accuracy than the supervised approach on both VLM and language benchmarks. Among the self-supervised methods, DINO achieved the highest accuracy of 50.95$\%$ on VLM benchmarks. Furthermore, the vision encoder trained with different paradigms showed minimal variation in language benchmarks, the accuracy difference between the best-performing SSL-DINO and the lowest-performing supervised learning approach was only 0.7$\%$. This also demonstrates that \textbf{embedding visual encoder pre-trained with various pre-training methods effectively preserves the native linguistic capabilities}.



\subsection{Elastic Visual Expert Based on ViT}
Our proposed Eve framework encompasses three training stages, strategically embedding elastic vision experts at each phase. In the initial two stages, we can leverage well-pretrained visual encoders from the public community, such as ResNet and ViT, which possess elasticity to construct robust visual capabilities. This section validates the performance of the ViT-based elastic vision experts in multimodal and linguistic abilities.

\subsubsection{Vision Expert Selection}

In the third stage, we integrate elastic visual experts into the framework, with a critical note that the language branch remained frozen. To investigate the influence of the visual expert's position on accuracy, we conducted experiments by fixing either the first or the second expert. As depicted in Table~\ref{fix_vision_order_new}, the first row corresponds to fixing the first expert while training the second expert, whereas the second row presents the opposite setup. The findings indicate that the difference between these two configurations is negligible. This suggests that \textbf{the position and order of the visual experts are not of paramount importance}. Moreover, when both experts participate in training and their responsibilities are balanced, the model exhibits the best performance on the both VLM and Language benchmarks. 
\begin{table*}[ht!]
 
	\begin{center}

        \scalebox{0.9}
            {
            \renewcommand{\arraystretch}{1.2}
    		\begin{tabular}{c|c|c|c|c|c|c|c|c}

    			\hline
       		   & & \multicolumn{3}{c|}{\bf{Image Question Answering}}     &\multicolumn{3}{c|}{\bf{Benchmark Toolkit}} & \ \\
    			\cline{3-8}
    			{\bf{Frist Expert}} &{\bf{Second Expert}} & {\bf{GQA}}& {\bf{SQA$^\mathrm{I}$}} & {\bf{VQA$^\mathrm{T}$}}& {\bf{POPE}} & {\bf{MME$^\mathrm{P}$}}& {\bf{MMB$^\mathrm{dev}$}}& {\bf{AVG}} \\
                & & 12578  & 2017 & 5000  & 8910  & 2374 & 4329 &  \\

    			\hline
                 Fixed & Trainable & 58.94  & 60.59 & 43.93 & 84.41 & 1261.57 & 57.04 & 61.33 \\
    		 Trainable &  Fixed & 59.41  & 59.63 & 44.23 & 84.66 & 1252.47  & 56.53 & 61.18 \\
    	   Trainable & Trainable & 59.83  & 60.24 & 44.61 & 84.8 & 1321.22  & 58.68 & \textbf{62.37}\\ 
                \hline
    		\end{tabular}}
	\end{center}
	\caption{Effect of fixing experts at different positions on the accuracy of multimodal tasks. "AVG" stands for the weighted mean of 6 VLM benchmarks.}
\setlength{\belowcaptionskip}{-0.1cm}
	\label{fix_vision_order_new}
\end{table*}

\subsubsection{Effect of Elastic Vision FFN}

To investigate the efficacy of the Elastic Vision FFN (EVF), we conducted experiments using pre-trained visual encoders based on ViT as well, with detailed results presented in Table ~\ref{EVE-layer_vit}. In our experimental setup, the visual branch was grounded in the CLIP ViT-L/14, while the linguistic branch was based on PanGu-$\pi$-1.5B, and the fusion module utilized LDP. The experimental results indicate that the use of the EVF layer led to an average accuracy improvement of 1$\%$ compared to the second phase models in multimodal tasks. This improvement can be attributed to the optimized token allocation strategy, which still resulted in a 0.7$\%$ improvement over the original MoE layer. Moreover, in language tasks, employing the MoE approach resulted in a 3$\%$ decrease in the accuracy of language tasks. In contrast, using the Eve model maintained the same level of accuracy as the second phase, thereby preserving the integrity of the language model to the greatest extent possible. 

\begin{table*}[ht!]
    
	\begin{center}
            \renewcommand{\arraystretch}{1.2}
            {
    		\begin{tabular}{ccccc}
    			\hline
    			{\bf{Stage}} & {\bf{EVF Layer}} & {\bf{MoE layer}} & {\bf{VLM AVG}} & {\bf{Lanuage AVG}}\\

    			\hline
                Stage2 &  &   & 61.23 &  58.65 \\
    	   Stage3 &  & \checkmark   &  \textbf{62.23} &  55.03 \\ 
    		 Stage3 & \checkmark &  &  61.93 & \textbf{58.65} \\
                \hline
    		\end{tabular}}
	\end{center}
    \setlength{\belowcaptionskip}{-0.3cm}
	\caption{Effect of the EVF layer or MoE layer on the average accuracy of VLM and language benchmarks.}
	\label{EVE-layer_vit}
\end{table*}

\subsubsection{Effect of Token Allocation Methods}


To evaluate the effectiveness of the EVF layer under different token allocation methods, we further examined the impact of three distinct token allocation strategies based on ViT. In our experiments, the visual branch employed CLIP ViT-L/14, the language branch used PanGu-$\pi$-1.5B, and the fusion module consisted of LDP. The experimental results are available in Table~\ref{token_sellection_acc_vit}. By implementing GBPR, the model's accuracy improved compared to the random allocation method, with an average increase of 0.4$\%$. Furthermore, the application of Img-GBPR resulted in a more substantial increase in accuracy, adding an additional 0.5$\%$ when compared to the random allocation method.

\begin{table*}[ht!]
	\begin{center}

        \scalebox{0.98}
            {
            \renewcommand{\arraystretch}{1.2}
    		\begin{tabular}{c|c|c|c|c|c|c|c}

    			\hline
       		  \multirow{2}{*}{\bf{Token Allocation}} & \multicolumn{3}{c|}{\bf{Image Question Answering}}     &\multicolumn{3}{c|}{\bf{Benchmark Toolkit}} & \multirow{2}{*}{\bf{AVG}}\ \\
    			\cline{2-7}
    			 & {\bf{GQA}}& {\bf{SQA$^\mathrm{I}$}} & {\bf{VQA$^\mathrm{T}$}}& {\bf{POPE}} & {\bf{MME$^\mathrm{P}$}}& {\bf{MMB$^\mathrm{dev}$}}&  \\

    			\hline
                Random & 59.75  & 59.00 & 44.52 & 84.92 & 1308.15 & 57.99 & 61.93 \\
    		 GBPR & 59.83  & 60.24 & 44.61 & 84.80 & 1321.22  & 58.68 & 62.37 \\
    	   Img-GBPR  & 59.08  & 61.87 & 44.50 & 84.26 & 1286.81  & 60.82 & \bf{62.48} \\ 
                \hline
    		\end{tabular}
      }
	\end{center}

    \setlength{\belowcaptionskip}{-0.3cm}
	\caption{Ablation study about different token allocation methods in Stage 3 with LLaVA-mixed-665k dataset.}
	\label{token_sellection_acc_vit}
\end{table*}

\subsection{SOTA Model Evaluation in Language Tasks}
We thoroughly evaluated the performance of the current state-of-the-art SVLMs in terms of language capability in Table~\ref{sota_result_inllm}. The experimental results indicate that the Eve model not only fully retains language capabilities but also significantly outperforms other small-sized (less than 3B) multimodal models in language tasks.

\begin{table*}[ht!]
	\begin{center}
        \renewcommand{\arraystretch}{1.2}
        \scalebox{0.88}
            {
    		\begin{tabular}{c|c|c|c|c|c|c|c|c|c|c|c|c}
    			\hline
    	
    			\multirow{2}{*}{\bf{Methold}} & \multirow{2}{*}{\bf{VM}} & \multirow{2}{*}{\bf{LLM}}  & \multirow{2}{*}{\bf{Res.}}  & \multicolumn{8}{c|}{\bf{Language Benchmarks}} & \multirow{2}{*}{\bf{AVG}} \\
    			\cline{5-12}
    	
    			 &  &   &  & {\bf{ceval}}& {\bf{cmmlu}} & {\bf{mmlu}}& {\bf{BoolQ}} & {\bf{Ax-b}} &{\bf{PIQA}}& {\bf{EPRSTMT}}& {\bf{XSum}}&  \\
    			\hline
    	   MobileVLM v2 1.7B &  ViT-L & M-1.4B & 336 &  27.50  & 25.88 & 31.64 & 56.02 & 46.9  & 66.97 & 46.25  & 19.75  & 40.12 \\ 
    	   MobileVLM v2 3B &  ViT-L & M-2.7B & 336 & 27.41 & 28.54 & 39.87 & 53.94 & 59.33  & 73.18 & 47.50 &20.55  & 43.79  \\
    	   LLaVA-Phi2.7B &  ViT-L & P-2.7B & 336 &  31.81  & 32.11 & 58.45 & 77.77 & 43.30 & 78.94 & 46.25  & 13.78  & 46.08 \\ 
        
             deepseek-vl & SigLIP & D-1B & 384 & 33.36 & 32.87 & 32.50 & 54.62 & 47.55 & 60.88 & 74.38 & 18.69  & 45.70  \\
    
                \hline
             Eve & ViT-L & PG-1.5B & 336 & 53.69 & 50.11 & 50.15  & 64.74  & 47.55 & 74.48 & 90.00 & 37.73 & \textbf{58.56} \\  
            
             \hline
    		\end{tabular}}
	\end{center}
        \setlength{\belowcaptionskip}{-0.3cm}
 	\caption{Comparison among different SVLMs on language benchmarks. “Res.” represents the input image resolution, “AVG" represents the mean of 8 language benchmarks.}
	\label{sota_result_inllm}
\end{table*}

\section{Dataset Details}
Since the training data used in the first two phases of this study are identical to the training data used in the first two phases of MobileVLM, this response will provide a comprehensive and detailed description of the unique training data utilized in the third phase. Table ~\ref{data_distribution_app} provides details on the data utilized during the training phase.

\begin{table}[ht!]
    \small

	\begin{center}
    \renewcommand{\arraystretch}{1.3}
        \fontsize{6}{6}\selectfont
        \setlength{\tabcolsep}{1mm}
            {
    		\begin{tabular}{l|l|l|c}
    			\hline
    			{\bf{Stage}} & \bf{Class} & {\bf{Data}}& {\bf{Samples}} \\
    			\hline
                Stage1 & LLaVA-1.5 & LLaVA-558k~\cite{liu2023improved}  & 558K \\
                \hline
                Stage2 & LLaVA-1.5 & LLaVA-mixed-665k ~\cite{liu2023improved} & 665K \\
                \hline
                \multirow{24}{*}{Stage3} & \multirow{3}{*}{General Multi-modality} & ShareGPT4V~\cite{chen2023sharegpt4v} & 665K \\
                 &  & LAION-GPTV~\cite{schuhmann2021laion400m} & 500K \\
                 &  & VFLAN~\cite{xu2024vision} & 254K \\
                 \cline{2-4}
                & \multirow{8}{*}{VQA} & ScienceQA~\cite{lu2022learn} & 12K \\
                & & IconQA~\cite{lu2021iconqa} & 107K \\
                & & TextVQA~\cite{singh2019towards} & 35K \\
                & & VSR~\cite{liu2023visual} & 13K \\
                & & OKVQA~\cite{marino2019ok} & 10K \\
                & & VQAV2~\cite{goyal2017making} & 10K \\
                & & KVQA~\cite{shah2019kvqa} & 24K \\
                & & TQA~\cite{kembhavi2017you} & 12K \\
                \cline{2-4}
                & \multirow{6}{*}{OCR} & AI2D~\cite{kembhavi2016diagram} & 4K \\
                & & DocVQA~\cite{tito2021document}& 10K \\
                & & ChartQA~\cite{masry2022chartqa} & 4K \\
                & & CTW~\cite{liu2019curved} & 1K \\
                & & DVQA~\cite{kafle2018dvqa} & 10K \\
                & & STVQA & 26K \\
                \cline{2-4}
                & \multirow{3}{*}{Caption} & SBU~\cite{ordonez2011im2text} & 843K \\
                & & COCO~\cite{chen2015microsoft} & 592K \\
                & & TextCaps~\cite{sidorov2020textcaps} & 16K \\
                \cline{2-4}
                & \multirow{3}{*}{Knowledge} & Landmark~\cite{weyand2020google} & 15K \\
                & & MovieNet~\cite{huang2020movienet} & 1K \\
                & & STEM~\cite{shen2024measuring} & 60K \\
                \cline{2-4}
                & Total &  & 3.2M \\

                 \hline
    		\end{tabular}}
	\end{center}
    
	\caption{Training datasets used for Eve.}
 \label{data_distribution_app}
\end{table}
\subsection{General Multi-modal Data}
\begin{figure*}
  \centering
 \includegraphics[width=0.98\linewidth] 
            {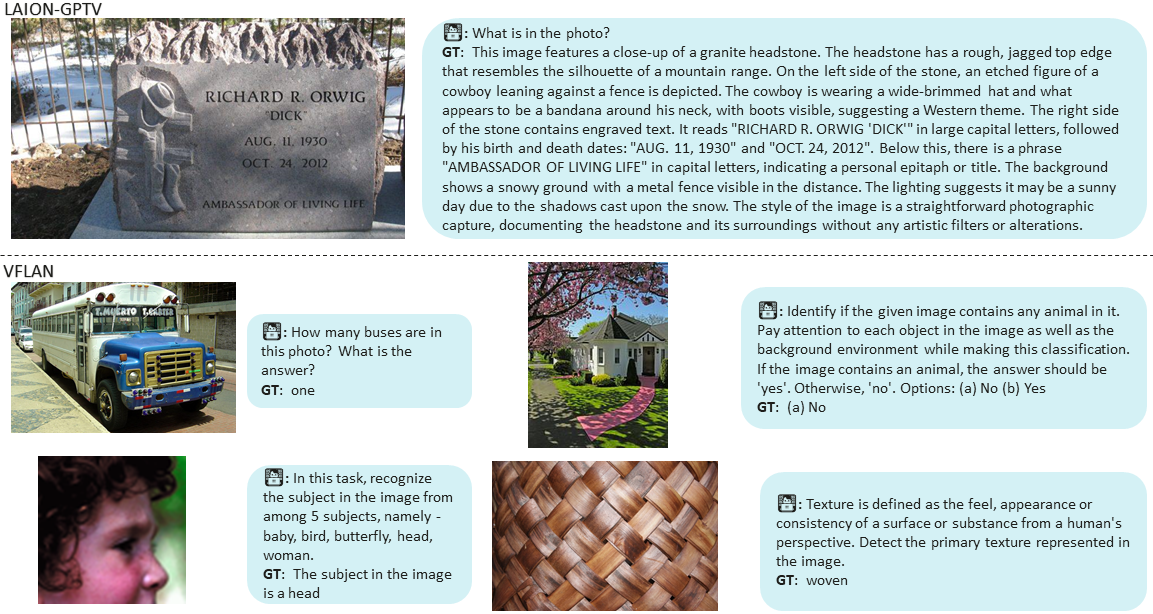} 

  \hfill
  \setlength{\abovecaptionskip}{-0.3cm}
  
  \setlength{\belowcaptionskip}{0.1cm}
  \caption{Example images of the general multi-modal dataset.}
  \label{fig:general}
\end{figure*}

ShareGPT4V is a large-scale, high-quality multimodal dataset built upon GPT4-Vision, which is frequently employed for training tasks during the instruction tuning phase of numerous multimodal models. For further details regarding the data, please refer to~\cite{chen2023sharegpt4v}. We extracted 500k and 254k training samples from LAION-GPTV and VFLAN, respectively, for our third-stage training tasks. The sample details are shown in Fig~\ref{fig:general}.

\subsection{VQA}
\begin{figure*}
  \centering
 \includegraphics[width=0.98\linewidth] 
            {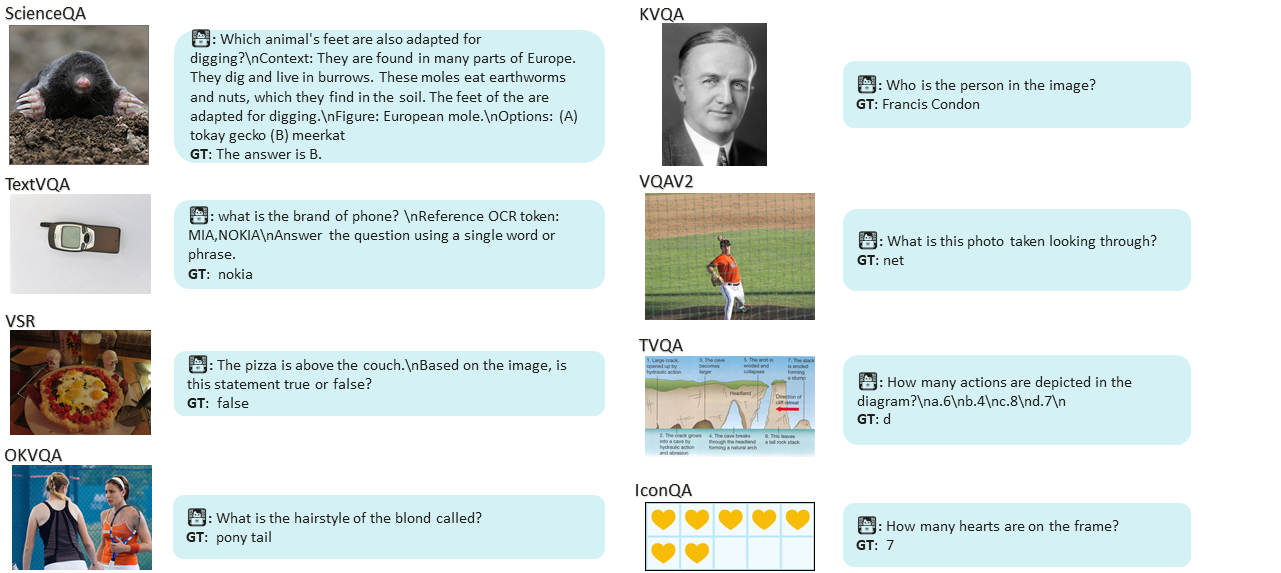} 

  \hfill
  \setlength{\abovecaptionskip}{-0.3cm}
  
  \setlength{\belowcaptionskip}{0.1cm}
  \caption{Display of examples of VQA data.}
  \label{fig:vqa}
\end{figure*}

The VQA datasets employed in the third stage are depicted in Fig~\ref{fig:vqa}. These datasets encompass ScienceQA, TextVQA, VSR, OKVQA, KVQA, VQAV2, TVQA and IconQA. Specifically, for the VQAV2 dataset, we opt for the answer exhibiting the highest confidence level. Conversely, no additional processing is conducted on the remaining datasets.
\begin{itemize}
  \item ScienceQA: In this study, we used a subset of 12,000 training samples from ScienceQA, a multimodal scientific question-answering dataset, which was incorporated into the training process of the third stage.
  \item TextVQA: In this research, we utilized a subset of 35,000 training instances from TextVQA, a dataset designed for text-based visual question answering, which was integrated into the training phase of the third stage.
  \item VSR: Visual Spatial Reasoning (VSR) is a dataset consisting of caption-image pairs with true/false labels. Each caption describes the spatial relationship between two objects in the image, and a Visual Language Model (VLM) is required to determine whether the caption accurately describes the image (true) or not (false). In this study, we used a selection of 13,000 training instances from the VSR dataset in the third stage of training.
  \item OKVQA: Unlike traditional VQA, Object-Driven Visual Question Answering (OKVQA) places a particular emphasis on understanding the objects and their attributes within an image. It requires the model to not only rely on a global scene understanding when answering questions but also to accurately identify and comprehend specific objects and their attributes in the image. In this study, we incorporated a selection of 10,000 training instances from the OKVQA dataset into the third stage of training.
  \item KVQA: Knowledge-based Visual Question Answering (KVQA) refers to Knowledge-based Visual Question Answering, which is a dataset emphasizing the need for models to utilize external knowledge bases or common sense for reasoning when answering questions related to images, rather than simply extracting the answer directly from the image content. This article selected 24k training samples from this dataset to be included in the training of the third stage.
  \item VQAV2: VQAv2 is an expansion and improvement over VQAv1, featuring increased diversity and complexity of questions. This paper incorporates 10k samples from the training set of VQAv2 into the training for the third phase.
  \item TVQA: Textbook Question Answering (TQA) dataset aims at answering multimodal questions given a context of text, diagrams and images. This paper incorporates 12k samples from the training set of TVQA into the training for the third phase.
  \item IconQA: IconQA is a dataset that involves understanding icons and visual question answering. Consistent with MobileVLM v2 in the second stage, this paper incorporates 107k IconQA data.
  
\end{itemize}

\subsection{OCR}

\begin{figure*}
  \centering
 \includegraphics[width=0.98\linewidth] 
            {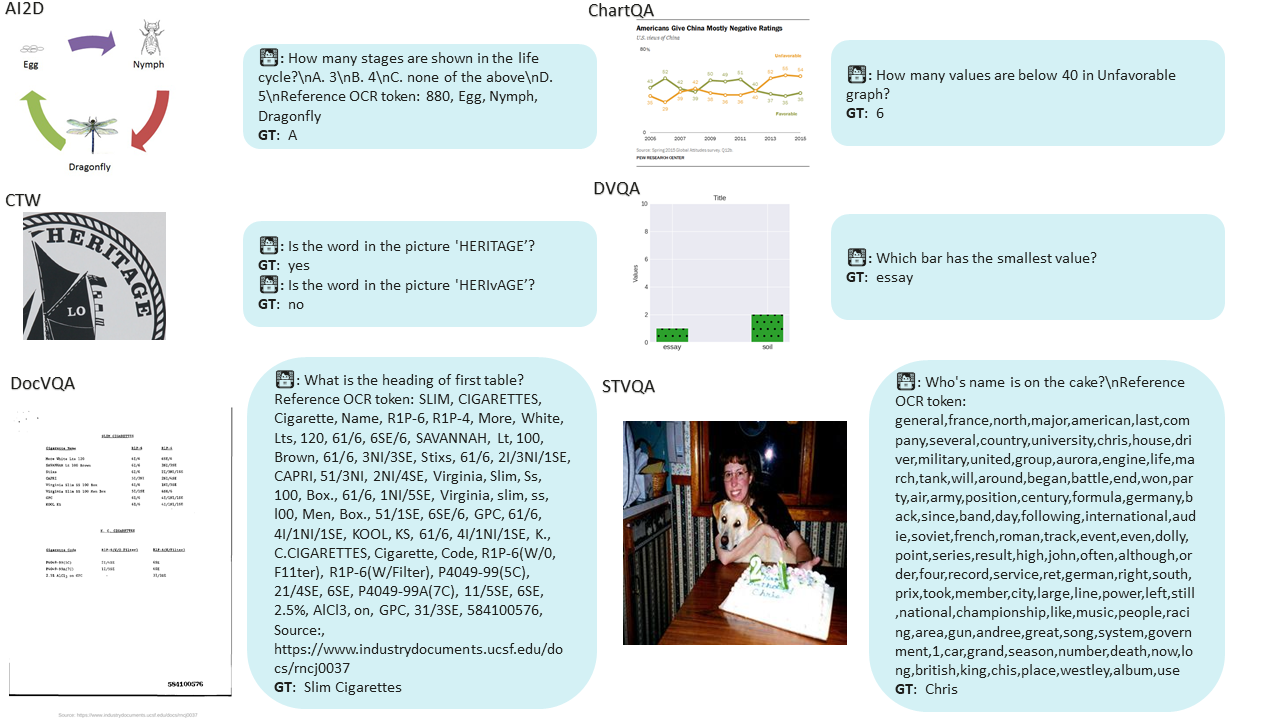} 

  \hfill
  \setlength{\abovecaptionskip}{-0.3cm}
  
  \setlength{\belowcaptionskip}{-0.5cm}
  \caption{Visualization of samples of OCR-related data.}
  \label{fig:ocr}
\end{figure*}

The OCR datasets utilized in the third stage are showcased in Fig~\ref{fig:ocr}. These datasets consist of AI2D, DocVQA, ChartQA, CTW, DVQA and STVQA. Notably, for the CTW dataset, we randomly enlarge the bounding boxes of the detected text to highlight the text elements that require recognition within the image. And for each text-image, generate two questions in the format "Is the word in the picture '{key-word}'?", where the positive samples have the key-word being the target word to be recognized in the image, while for negative samples, we will randomly replace, add, or delete one or more letters from the key-word. For DocVQA and STVQA, OCR tokens will be used as the hint information for the questions.
\begin{itemize}
  \item AI2D: Artificial Intelligence for Image Description (AI2D), specifically, addresses the intricate task of interpreting diagrams and illustrations within the realm of educational materials. To further enhance the complexity and applicability in our research, we integrate a comprehensive 4k AI2D training set into the training process during the third stage. 
  \item DocVQA: Document Visual Question Answering (DocVQA) involves understanding and answering questions based on visual content found in documents, such as forms, receipts, tables, charts, and other types of document images. For the third stage of our training process, we have incorporated a carefully selected 10k dataset to enrich and enhance the overall performance and applicability of the DocVQA dataset.
  \item ChartQA: Chart-based Visual Question Answering (ChartQA) focuses on interpreting charts such as bar graphs, pie charts, line graphs, scatter plots, and other data visualization formats commonly used in reports, presentations, and scientific publications. this paper incorporates 4k ChartQA data.
  \item CTW: The Curved Text Detection in the Wild (CTW) dataset focuses on the detection and recognition of curved or irregularly shaped text in natural scene images. To enhance the OCR recognition capabilities of the model, this paper randomly selected 1,000 training data points and incorporated them into the third phase of the task.
  \item DVQA: The Density Visual Question Answering (DVQA) dataset not only requires recognizing objects or scenes but also involves making quantitative assessments and comparisons. In this paper, 10k training data were added to the third stage.
  \item STVQA: In Scene Text Visual Question Answering (STVQA) tasks, the model must not only recognize and read the text within the image accurately but also understand the context in which the text appears and how it relates to the question being asked. In this paper, 26k training data were added to the third stage.

\end{itemize}
\begin{figure*}
  \centering
 \includegraphics[width=0.98\linewidth] 
            {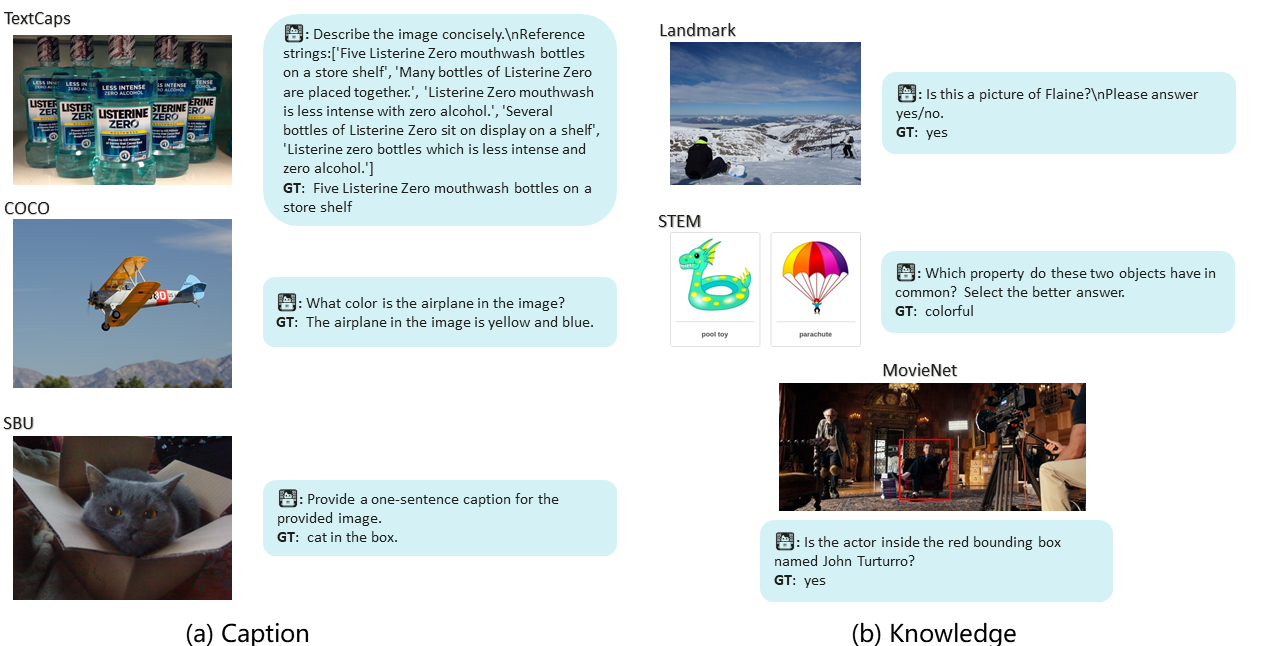} 

  \hfill
  \setlength{\abovecaptionskip}{-0.3cm}
  
  \setlength{\belowcaptionskip}{-0.5cm}
  \caption{Visualization of samples of caption and knowledge data.}
  \label{fig:caption}
\end{figure*}

\subsection{Caption}
The caption dataset utilized in the third stage, as depicted in Fig~\ref{fig:caption}.a, comprises 843k SBU, 592k COCO, and 16k TextCaps. Notably, these datasets align with the data employed in MobileVLM v2.

\subsection{Knowledge}
The knowledge dataset employed in the third stage is depicted in Fig~\ref{fig:caption}.b, which encompasses Landmark, MovieNet, and STEM. For the Landmark and MovieNet datasets, we have crafted two sets of questions each—one set targeting correct responses and another set with incorrect responses. Specifically for MovieNet, we have highlighted the movie tasks that require accurate identification with red boxes, based on the annotation information.
\begin{itemize}
  \item Landmark: In our third stage of training, we have generated 15,000 question-and-answer pairs utilizing the Google Landmarks dataset (GLDv2). This dataset comprises images meticulously annotated with human-made and natural landmark labels, providing a rich resource for our question-and-answer development.
  \item MovieNet: MovieNet encompasses a comprehensive dataset featuring 1.1 million characters delineated by bounding boxes and identified by unique identities, 42,000 scene boundaries, 2,500 aligned description sentences, 65,000 tags categorizing places and actions, and 92,000 tags specifying cinematic styles. Leveraging this extensive resource, we have crafted 1,000 question-and-answer pairs centered around movie characters for the third phase of our project.
  \item STEM: The STEM (Science, Technology, Engineering, and Mathematics) domain encompasses a vast array of interconnected fields and disciplines. In this article, we have elected to focus on the 'S' in STEM, extracting 60k pertinent training data related to the science field and incorporating it into our study's training phase. This strategic approach allows us to delve deeper into the scientific aspects of STEM, thereby enhancing the overall depth and breadth of our research.
  
\end{itemize}

\section{Broader Impacts and Limitations}
\paragraph{Broader Impacts.}
In the case of Eve, the training dataset is largely compiled from the vast expanse of the internet. This inherently exposes the model to a broad spectrum of information, encompassing diverse viewpoints and representations. However, this openness also poses challenges, as the internet content is not uniformly balanced or unbiased. There exists a myriad of perspectives online, some of which can be skewed, misinformed, or reflective of societal prejudices. During the training phase, Eve learns by identifying patterns and correlations within this data. Unfortunately, this process can inadvertently lead to the absorption and amplification of imbalances present in the training data. Biases in the form of disproportionate representation or stereotyping that exist within the source material may thus be inadvertently encoded into the model's learned parameters. Consequently, when Eve generates outputs, there is a risk that these biases and discriminatory elements, absorbed from the unfiltered internet content, may manifest in its responses. This could range from perpetuating gender or racial stereotypes in language generation to reflecting and reinforcing socio-political biases present in the training dataset. We are committed to rigorously screening for biased data and endeavor to prevent the application of our model for political or military purposes.

\paragraph{Limitations.}
In our investigation of Small Visual Language Models (SVLMs), we have discerned not only a positive correlation between the linguistic competence of these models and their multi-modal prowess, but also the distinct influence of training data types on their performance. This variability restricts the generalizability of training data for multi-modal processing, emphasizing the need for caution in applying our findings to newly emerging language models. It is crucial to recognize that the effectiveness of our training datasets may not be universally transferrable, necessitating further validation and tailored selection. To enhance the overall capabilities of multi-modal models, these efforts should complement the inherent strengths of the language models themselves.
\clearpage
\bibliography{aaai25}
\end{document}